\documentclass[10pt,conference,anonymous]{IEEEtran}
\IEEEoverridecommandlockouts

\usepackage{cite}
\usepackage{amsmath,amssymb,amsfonts}
\usepackage{algorithmic}
\usepackage{graphicx}
\usepackage{pifont} 
\usepackage{textcomp}
\usepackage{xcolor}
\usepackage{multirow}
\usepackage{tcolorbox}
\usepackage{colortbl}
\usepackage{geometry}
\usepackage{booktabs}
\usepackage{pgfplots}                                                     \pgfplotsset{compat=1.18}
\def\BibTeX{{\rm B\kern-.05em{\sc i\kern-.025em b}\kern-.08em
    T\kern-.1667em\lower.7ex\hbox{E}\kern-.125emX}}
\begin{document}

\newtcolorbox{summarybox}{
    enhanced,                     
    boxrule=0pt,                  
    frame hidden,                 
    borderline west={4pt}{0pt}{darkgray}, 
    colback=gray!12,              
    sharp corners,                
    left=8pt,                     
    right=8pt,                    
    top=6pt,                      
    bottom=6pt,                   
    before skip=10pt,             
    after skip=10pt               
}

\title{Hawk: Harnessing Hardware-Aware Knowledge for High-Performance NPU Kernel Generation
}

\author{
\IEEEauthorblockN{
Junyi Wen\IEEEauthorrefmark{1},
Ruiyan Zhuang\IEEEauthorrefmark{2},
Yongjia Xu\IEEEauthorrefmark{1},
Pengtu Li\IEEEauthorrefmark{1},
Rui Zou\IEEEauthorrefmark{1}, \\
Hongyi Chen\IEEEauthorrefmark{1},
Chingman Wan\IEEEauthorrefmark{1},
Puxu Yang\IEEEauthorrefmark{1},
Wuhui Chen\IEEEauthorrefmark{1}\IEEEauthorrefmark{3}, and
Yanlin Wang\IEEEauthorrefmark{1}
}

\vspace{0.2cm}

\IEEEauthorblockA{\IEEEauthorrefmark{1}\textit{Sun Yat-sen University}, Zhuhai, China \\
Email: \{wenjy23, xuyj73, lipt5, zour5, chenhy567, wancm5, yangpx26\}@mail2.sysu.edu.cn, \\
\{chenwuh, wangylin36\}@mail.sysu.edu.cn}

\vspace{0.1cm}

\IEEEauthorblockA{\IEEEauthorrefmark{2}\textit{Greater Bay Area National Technology Innovation Center}, Guangzhou, China \\
Email: zhuangruiyan@ncti-gba.cn}

\vspace{0.1cm}

\IEEEauthorblockA{\IEEEauthorrefmark{3}\textit{Peng Cheng Laboratory}, Shenzhen, China}
}

\maketitle

\begin{abstract}
Developing high-performance kernels for Neural Processing Units (NPUs) is a critical industry bottleneck, requiring developers to manually navigate implicit hardware constraints and strict memory hierarchies. While large language models offer immense automation potential, they fail catastrophically on NPUs due to a fundamental lack of hardware-specific priors. Naïvely transplanting code snippets from similar NPU kernels may pass the compiler, but it consistently triggers runtime crashes and performance degradation by blindly violating underlying hardware constraints. To overcome this, we introduce \textbf{Hawk}, a training-free framework that harnesses hardware-aware knowledge through three core modules: (1) \textbf{Run-Time Knowledge Synthesis Module}, which employs a \textit{Triple-Part Executable Knowledge Representation} to inherently couple the error context with executable semantics; (2) \textbf{Bottleneck-Aware Knowledge Retrieval Module}, which implements a \textit{2D-Retrieval} paradigm to project queries into orthogonal syntactic and hardware-aligned semantic spaces; and (3) \textbf{Effect-Driven Knowledge Distillation Module}, which leverages \textit{LLM-driven semantic arbitration} to continuously distill the knowledge by pruning errors and consolidating redundancies based on the empirical execution feedback. Extensive evaluations on real-world NPU workloads demonstrate that Hawk elevates generation accuracy from 49.4\% to 80.0\%, while achieving up to a 2.2$\times$ execution speedup over state-of-the-art baselines.
\end{abstract}

\begin{IEEEkeywords}
NPU, Kernel Generation,  Harness Engineering, Automatic Programming.
\end{IEEEkeywords}

\vspace{-5pt}

\section{Introduction}

Domain-specific accelerators like Neural Processing Units (NPUs) are critical for modern AI workloads, yet automating NPU kernel generation via Large Language Models (LLMs)~\cite{xu2026deepseek, team2026qwen3, zeng2026glm, achiam2023gpt} remains notoriously difficult. Our empirical study (\S \ref{motivation}) shows state-of-the-art LLMs achieve merely 13.3\% Comp@1 (compilation success rate) and 13.3\% Pass@1 (functional correctness rate) on NPUs, starkly contrasting with GPU generation (100\% Comp@1, 80\% Pass@1). We find that over 75\% of these failures stem from low-level, hardware-specific misuses (e.g., invalid API invocations and memory layout mismatches). Moreover, even functionally correct NPU kernels execute nearly 20$\times$ slower than their GPU equivalents under identical theoretical compute capacity.

Recent efforts to improve NPU kernel generation generally fall into two paradigms: model training (e.g., domain-adaptive fine-tuning~\cite{cao2026ascendkernelgen} and RL guidance~\cite{zheng2026towards}) and intermediate representation (IR) assistance (e.g., lightweight DSLs~\cite{wen2026ascendcraft} and polyhedral compilers~\cite{zhao2021akg}). However, both suffer from severe scalability bottlenecks. In model-centric approaches, API updates~\cite{mai2025towards, liu2022morest} induce distribution shifts that invalidate existing datasets, necessitating thousands of new expert-labeled samples and resource-intensive retraining~\cite{kernelllm2025, GargReinforcementTO, woo2025tritonrl}. Conversely, in IR-centric methods, even modest code library extensions demand months of manual engineering from specialists to rewrite grammar rules~\cite{zhan2024react, zhou2025type, shir2025robust, lagouvardos2025incredible}.

To bypass these scalability bottlenecks, we explore a highly scalable alternative: \textit{exploiting functional similarity among existing NPU kernels}. Our analysis (\S \ref{motivation}) reveals up to 70\% code similarity across open-source NPU kernels, suggesting immense potential for knowledge transfer. However, naive code transplantation~\cite{liu2025few} is brittle; functional similarity alone fails to capture the underlying hardware-aware knowledge. Overlooking such knowledge often triggers runtime crashes and severe performance degradation, despite successful compilation.

To this end, we propose Hawk, an LLM-based framework that \underline{h}arnesses hardware-\underline{aw}are \underline{k}nowledge for high-performance NPU kernel generation. Grounded in in-context learning theory~\cite{wu2025context, hahn2023theory}, Hawk maintains a curated knowledge base of existing NPU kernels. By dynamically retrieving relevant knowledge, it guides LLMs to avoid compilation and performance pitfalls. However, realizing this framework requires addressing three fundamental challenges:

\textbf{Challenge 1: Heterogeneous Knowledge Structuring. } Code snippets lack crucial optimization rationales and hardware constraints. Effective hardware-aware knowledge inherently spans natural language strategies, pseudo-code, and executable implementations. Directly concatenating them exhausts the LLM context window and dilutes critical signals, whereas forcing a uniform format destroys syntactic precision. The challenge is organizing this heterogeneous knowledge to preserve both semantic richness and syntactic fidelity.

\textbf{Challenge 2: Hardware-Aware Knowledge Retrieval.} Naive retrieval relies functional or textual code similarity, which is fundamentally inadequate for NPU kernels. Two kernels executing identical mathematical operations may require entirely different hardware implementations due to distinct memory layouts, tensor shapes, or hardware limits. The challenge lies in designing a retrieval mechanism that goes beyond superficial code similarity to accurately capture and match deep hardware-aware constraints.

\textbf{Challenge 3: Context-Aware Knowledge Maintenance.} Over time, the knowledge base inevitably accumulates erroneous and redundant entries. Erroneous knowledge directly misleads the LLM into generating invalid or crashing kernels. Redundant entries squander the critical context window budget, trapping the LLM in a loop of repeatedly fetching identical strategies. The challenge is to automatically identify and prune this low-quality knowledge based on empirical execution feedback without disrupting valid contextual dependencies.

Hawk incorporates the following three core modules:

\textbf{Run-Time Knowledge Synthesis (Addressing C1).} To achieve this synthesis without token exhaustion, this module introduces a Triple-Part Executable Knowledge Representation. Rather than flatly concatenating raw manuals and code, we deconstruct NPU domain knowledge into a strict informational hierarchy: (1) \textit{indexing triggers} (e.g., operator types or data types) for rapid context localization; (2) \textit{semantic rationales} (e.g., natural language rules regarding NPU Unified Buffer limits or tiling strategies) to explain the underlying hardware constraints and how to apply the knowledge; and (3) \textit{exact syntactic templates} (e.g., verified API code snippets) for executable deployment. Crucially, the knowledge is dynamically captured at run-time: whenever the agent meets predefined triggering conditions during kernel generation, it systematically formalizes newly discovered insights into this exact three-part format, ensuring the knowledge base continuously evolves.

\textbf{Bottleneck-Aware Knowledge Retrieval (Addressing C2).} To overcome the limits of naive similarity, this module implements a 2D-Retrieval paradigm. Moving beyond superficial symptom-based text matching, Hawk dynamically projects queries into two orthogonal dimensions: a syntactic space ensuring exact API signature compliance, and a hardware-aligned semantic space capturing deep memory and parallelization constraints. By fusing these dimensions, the module precisely isolates actionable, root-cause optimization strategies rather than merely fetching functionally analogous code snippets.

\textbf{Effect-Driven Knowledge Distillation (Addressing C3).} To maintain a high-purity corpus, this module employs an LLM-driven semantic arbitration. Rather than relying on static heuristic filtering, it actively leverages physical NPU compilation and execution feedback to continuously validate entries. It systematically purges context-poisoning errors and prunes functionally redundant patterns, ensuring the agent is guided solely by empirically verified, high-utility hardware constraints.

Our main contributions are:

\begin{itemize}
\item We systematically demonstrate that the absence of implicit hardware-aware knowledge is the fundamental bottleneck for LLMs to generate high-performance NPU kernels.
\item We propose \textbf{Hawk}, a training-free framework for high-performance NPU kernel generation with harnessing hardware-aware knowledge. By orchestrating Run-Time knowledge synthesis, bottleneck-aware retrieval, and effect-driven distillation, Hawk enforces NPU kernel generation without the exorbitant overhead of model fine-tuning.
\item Extensive evaluations on real-world Ascend NPU workloads prove that Hawk elevates the accuracy from 49.4\% to 80.0\% and delivers up to a 2.2$\times$ execution speedup compared to state-of-the-art works.
\end{itemize}


\section{Background and Related Work}

\begin{figure}[t]
    \centering
    \setlength\abovecaptionskip{0.02\baselineskip}
    \setlength\belowcaptionskip{-0.5\baselineskip}
	\includegraphics[width=0.8\linewidth]{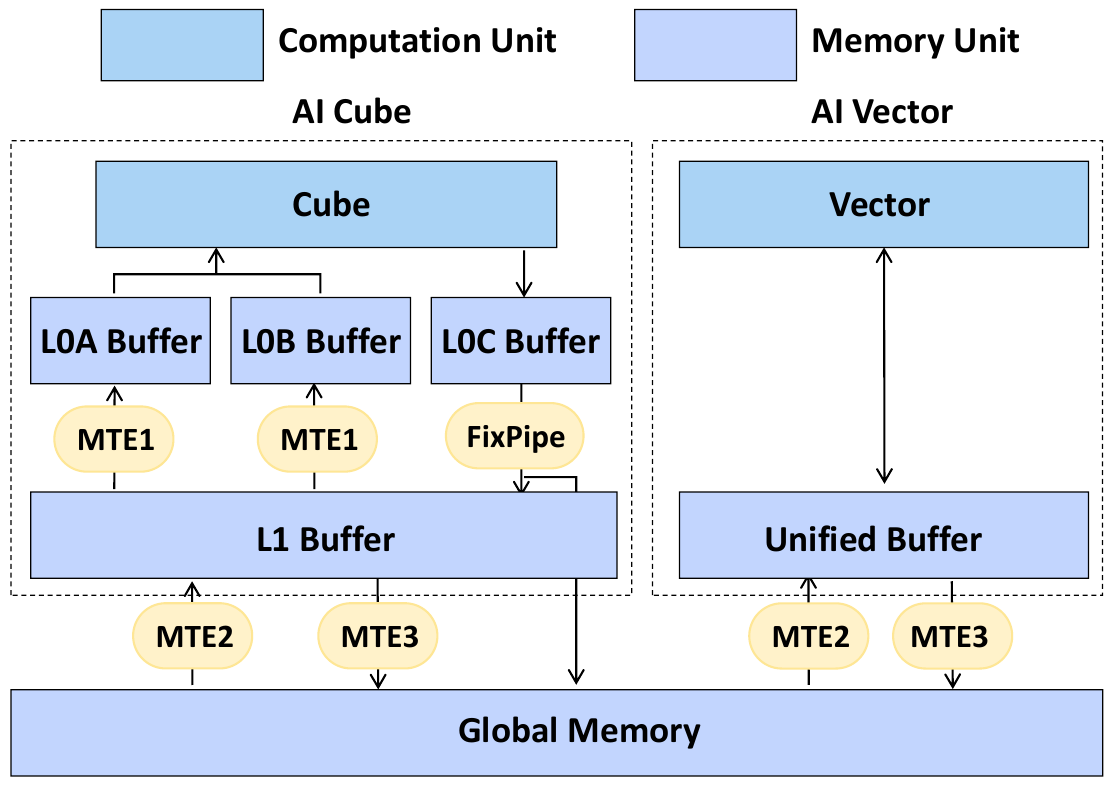}
	\caption{Ascend NPU architecture.}
    \label{NPU_architecture}
    \vspace{-15pt}
\end{figure}

This section contextualizes Hawk by introducing the NPU architecture, its programming model, and the limitations of current LLM-based generation.

\vspace{-8pt}

\subsection{NPU Architecture}

As shown in Figure \ref{NPU_architecture}, NPU employs a heterogeneous computing architecture~\cite{fernando2025edge, fang2026heterogeneous} designed to efficiently accelerate AI workloads. Its complexity and high performance stem from three tightly coupled hardware designs:

\textbf{Compute Units.} The computational core comprises two specialized units: the \textit{AI Cube}, optimized for dense matrix multiplications (completing a full $M \times K$ and $K \times N$ matrix multiplication in a single execution cycle), and the \textit{AI Vector}, dedicated to element-wise operations (e.g., ReLU, pooling).

\textbf{Memory Hierarchy \& Data Movement.} Ascend features a multi-tiered memory hierarchy (Global Memory, L1, Unified Buffer, and L0) that is \textit{fully exposed to software}, requiring developers to explicitly manage data allocation. Data migration~\cite{zhang2025skybyte, wang2025faster} across these tiers is orchestrated by multiple Memory Transfer Engines (MTEs). Operations within a single MTE are sequential, whereas across different MTEs in parallel, demanding meticulous coordination to maximize bandwidth.

\textbf{Explicit Instruction Pipelining.} To overlap computation with memory access, Ascend encapsulates tasks into instructions dispatched to independent queues (Scalar, Vector, Cube, and MTE). While intra-queue instructions run sequentially, cross-queue instructions execute concurrently. This enables fine-grained parallelism for NPU.


\subsection{LLM-based Kernel Generation}

Recent advancements in LLM-based kernel generation face a fundamental trade-off between domain adaptation costs and engineering agility. They can be broadly categorized into fine-tuning approaches and training-free, IR-Assisted frameworks.

\begin{table}[t]
\centering
\caption{Results under the add-shot strategy on 15 activation operators across two platforms and two models.}
\label{tab:add_shot_comp}
\small
\begin{tabular}{lcccc}
\hline
\textbf{Model} & \multicolumn{2}{c}{\textbf{CUDA}} & \multicolumn{2}{c}{\textbf{Ascend C}} \\
\cline{2-5}
 & Comp@1 & Pass@1 & Comp@1 & Pass@1 \\
\hline
DeepSeek-V4 & 100.0\% & 80.0\% & 13.3\% & 13.3\% \\
GLM-4.5 & 100.0\% & 86.7\% & 33.3\% & 20.0\% \\
\hline
\end{tabular}
\vspace{-15pt}
\end{table}

\textbf{Fine-Tuning Approaches.} Extensive efforts have leveraged Supervised Fine-Tuning (SFT) and Reinforcement Learning (RL) to specialize LLMs for GPU kernels. For instance, CUDA Agent~\cite{dai2026cuda} and Kevin~\cite{baronio2025kevin} construct massive, paired datasets of algorithmic descriptions and CUDA implementations for SFT. To further align models with hardware execution, CUDA-L1~\cite{li2025cuda} and KernelBlaster~\cite{dong2026kernelblaster} utilize automated compiler feedback and profiling metrics as reward signals to drive RL optimization. Recent literature adapts this parameter-updating paradigm to NPUs: AscendKernelGen~\cite{cao2026ascendkernelgen} fine-tunes base models on proprietary Ascend C corpora, while EvoKernel~\cite{zheng2026towards} employs evolutionary algorithms via RL to discover valid kernel variants. 

\begin{table}[t]
\centering
\caption{Classification of 126 failures.}
\label{tab:error_classification}
\begin{tabular}{p{3cm}ccc}
\hline
\textbf{Error Type} & \textbf{Count} & \textbf{Level} & \textbf{Prop.} \\
\hline
Non-existent function. & 27 & Syntax & -- \\
Incorrect type name & 5 & Syntax & -- \\
Host-side API misuse & 12 & Syntax & -- \\
Code formatting & 1 & Syntax & -- \\
Runtime timeout & 53 & Syntax & -- \\
\textbf{Syntax-level subtotal} & \textbf{98} & & \textbf{77.8\%} \\
\hline
UB overflow & 26 & Hardware & -- \\
\textbf{Hardware-level subtotal} & \textbf{26} & & \textbf{20.6\%} \\
\hline
Correctness mismatch & 2 & Numeric & -- \\
\textbf{Numeric-level subtotal} & \textbf{2} & & \textbf{1.6\%} \\
\hline
\textbf{Total} & \textbf{126} & & \textbf{100\%} \\
\hline
\end{tabular}
\vspace{-15pt}
\end{table}

While effective, fine-tuning faces severe scalability bottlenecks: continuously evolving hardware and Ascend toolchains necessitate costly dataset reconstruction and resource-intensive retraining to prevent performance degradation. In contrast, Hawk adapts instantaneously. By simply appending new expertise to its external knowledge base, Hawk seamlessly integrates hardware updates while entirely bypassing the exorbitant cost of retraining LLMs.

   \begin{table*}[t]
    \centering
    \setlength\abovecaptionskip{0.02\baselineskip}
    \scriptsize
    \setlength\belowcaptionskip{-0.5\baselineskip}
    \caption{The eight development-paradigm dimensions, their meanings, and
    coverage across 431 kernels.} 
    \label{tab:dim_def}                            
    \setlength{\tabcolsep}{3pt}
    \begin{tabular}{llll}
      \toprule                                         
      Functional dimension & Meaning & Representative APIs  \\                                                                       
      \midrule                                               
      Memory / Tensor   & Tensor abstractions and the on-chip memory pipeline every kernel relies on & GlobalTensor, TPipe, TQue  \\                     
      Data Movement   &  Data transfer between Global/UB/L1 memories---the I/O skeleton of a kernel             & DataCopy, CopyIn, CopyOut   \\
      Vector Compute  & Element-wise operations on the Vector unit (arithmetic, cast, compare)             & Cast, Add, Mul            \\
      Reduction       & Sum/max reductions along a dimension (means, variances, softmax denominators)           & ReduceSum, ReduceMax        \\
      Sync / Multicore  & Multi-core partitioning and pipeline event synchronization       & GetBlockIdx, PipeBarrier    \\                    
      Tiling Strategy & Tiling-driven block partitioning and dtype/algorithm dispatch      & GET\_TILING\_DATA            \\
      Cube / Matmul     & Cube-unit matrix-multiply APIs (matmul-family operators only)      & MatmulImpl, IterateAll      \\
      MicroAPI / Reg    & Register-level MicroAPI available only on the newest architecture               & MaskReg, RegTensor       \\
      \bottomrule                                                                                         
    \end{tabular}
    \vspace{-15pt}
  \end{table*}   

\textbf{Training-Free \& IR-Assisted Frameworks.} To avoid continuous retraining overhead, an alternative line of research focuses on multi-agent systems (MAS) and intermediate representations (IR). For mature ecosystems like GPUs, frameworks such as STARK~\cite{dong2025stark} and CudaForge~\cite{zhang2025cudaforge} orchestrate developer and evaluator agents to iteratively refine CUDA code using raw AST parsing and execution-driven self-reflection. Similarly, AVO~\cite{chen2026avo} integrates external memory to track historical optimization trajectories. However, these pure multi-agent systems intrinsically assume that the underlying LLM possesses deep pre-trained priors for the target language. Consequently, they falter completely on NPUs (\S\ref{motivation}), where the relied LLMs lack foundational NPU coding knowledge. 

To bridge this capability gap without fine-tuning, NPU-focused systems historically resort to IRs or Domain-Specific Languages (DSLs). For example, AKG~\cite{zhao2021akg} relies on polyhedral models for automatic scheduling, while AscendCraft~\cite{wen2026ascendcraft} builds transpilers that map high-level Python logic down to low-level Ascend C APIs. While effective, IR-centric pipelines demand exorbitant manual engineering: even trivial API updates force experts to rewrite grammars, transpilers, and AST rules. Instead of hardcoding brittle compiler logic, Hawk utilizes harnessing hardware-aware knowledge,  cleanly bypassing complex compiler engineering, coupling the agility of an LLM agent with strict domain knowledge.

\vspace{-8pt}

\section{Motivation}

\label{motivation}

This section reveals the fundamental limitations of LLMs in NPU kernel generation—typically authored in Ascend C—and the inadequacy of naive knowledge transfer, which motivates the design of Hawk, our hardware-aware knowledge harnessing framework.

\vspace{-5pt}

\subsection{Findings in NPU kernel generation with LLMs}
\label{subsec:NPUfinding}

\noindent\fcolorbox{black}{gray!8}{\parbox{\dimexpr\columnwidth-2\fboxsep-2\fboxrule\relax}{%
\textbf{Finding 1.} \textit{Despite strong CUDA code generation, SOTA LLMs struggle with Ascend C kernel generation.}
}}

To characterize LLM proficiency in NPU programming, we evaluate two state-of-the-art LLMs (DeepSeek-V4 Pro and GLM-4.5) on 15 activation kernels from MultiKernelBench~\cite{wen2025multikernelbench} using \textit{add-shot} prompting (providing a single vector-addition kernel as an example to help LLM). As Table~\ref{tab:add_shot_comp} shows, the performance gap is staggering. While the models achieve up to 100.0\% Comp@1 and 86.7\% Pass@1 on CUDA, Ascend C performance plummets. DeepSeek-V4 Pro yields only 13.3\% for both Comp@1 and Pass@1, and GLM-4.5 reaches merely 33.3\% Comp@1 and 20.0\% Pass@1.

Although both CUDA and Ascend C are C-based APIs, LLMs fail entirely to transfer their GPU programming capabilities to NPUs. To uncover the root cause of this severe degradation, we analyze the compilation logs to answer a critical question: \textit{What specific errors dominate these Ascend C generation failures?}

\noindent\fcolorbox{black}{gray!8}{\parbox{\dimexpr\columnwidth-2\fboxsep-2\fboxrule\relax}{%
\textbf{Finding 2.} \textit{The lack of API knowledge overwhelmingly dominates Ascend C generation failures.}
}}

To investigate the root causes of these failures, we manually classified all 126 errors from our trials into three categories: (1) \emph{Syntax-level errors} (violating Ascend C API specifications, e.g., hallucinated functions, type mismatches, or Host-side API misuse); (2) \emph{Hardware-level errors} (compilable code that violates physical constraints at runtime, e.g., exceeding Unified Buffer capacity); and (3) \emph{Numeric-level errors} (executable code yielding incorrect mathematical outputs).

\begin{figure}[t]
    \centering
    \setlength\abovecaptionskip{0.5\baselineskip}
    \setlength\belowcaptionskip{-0.5\baselineskip}
    \begin{minipage}[t]{0.48\columnwidth}
      \centering
      \includegraphics[width=\linewidth]{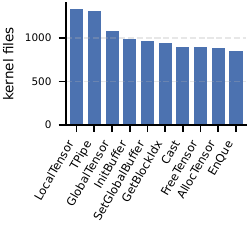}\\
      \caption{Top-10 API reference frequency across 2,690 kernel files.}
      \label{fig:api_frequency}
    \end{minipage}\hfill
    \begin{minipage}[t]{0.48\columnwidth}
      \centering
      \includegraphics[width=\linewidth]{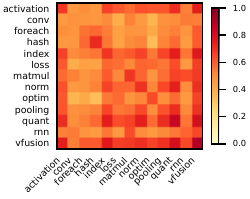}\\
      \caption{The functional dimension Jaccard of kernels.}
      \label{fig:cat_jaccard}
    \end{minipage}
    \vspace{-15pt}
\end{figure}

As shown in Table~\ref{tab:error_classification}, syntax-level errors dominate, accounting for 77.8\% of all failures. This rampant API hallucination and misuse indicates that LLMs fundamentally lack the hardware-specific syntactic priors required for Ascend C.

These severe syntax deficits motivate a highly scalable alternative: transferring API usage patterns from existing, correct NPU kernels to guide the LLM during generation. However, this knowledge transfer is only viable if NPU kernels inherently share substantial structural and API-level similarities. Thus, we assess the degree of cross-kernel commonality in open-source implementations to validate the feasibility of this approach.

\vspace{1pt}
\noindent\fcolorbox{black}{gray!8}{\parbox{\dimexpr\columnwidth-2\fboxsep-2\fboxrule\relax}{%
\textbf{Finding 3.} \textit{NPU kernels exhibit extensive overlap in both fine-grained API usage and coarse-grained functional paradigms, confirming the immense potential for cross-kernel syntax and structural transfer.}
}}

We analyze \textit{ops-nn}~\cite{opsnn}, a real-world code library comprising 2,690 NPU kernel files, assessing similarity across two dimensions:

\textbf{Fine-Grained API Reuse.} Figure \ref{fig:api_frequency} illustrates the per-API reference frequency. The most accessed API (LocalTensor) appears in 49.5\% of files, with the top-10 APIs all exceeding 30\%, indicating substantial potential for exact-match syntax transfer.

\textbf{Coarse-Grained Paradigm Similarity.} Since strict API matching may underestimate shared principles, we abstract all APIs into eight functional categories (Table \ref{tab:dim_def}). We represent each kernel as a mathematical set of these categories and calculate pairwise similarity using the Jaccard index. As shown in Figure \ref{fig:cat_jaccard}, nearly all similarities exceed 0.5, with many surpassing 0.7. This confirms that even when specific APIs differ, the underlying structural paradigms remain highly consistent.

\begin{figure}[t]
    \centering
    \setlength\abovecaptionskip{0.02\baselineskip}
    \setlength\belowcaptionskip{-0.5\baselineskip}
    \includegraphics[width=\linewidth]{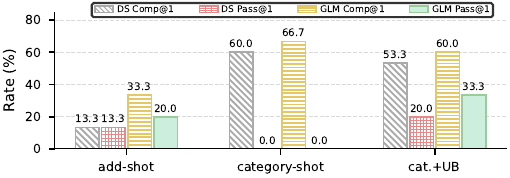}
    \caption{Effect of prompting strategy on Ascend C generation. DS = DeepSeek-V4 Pro, GLM = GLM-4.5. Pass@1 drops to 0\% with category-shot across all 15 kernels.}
    \label{fig:prompt_strategy}
    \vspace{-20pt}
\end{figure}

Driven by these empirical signals, we hypothesize that transferring these shared API paradigms serves as a viable foundational step for kernel generation. We next explore how to operationalize this syntax transfer, and critically, whether syntax alone is sufficient for successful execution.

\vspace{4pt}

\noindent\fcolorbox{black}{gray!8}{\parbox{\dimexpr\columnwidth-2\fboxsep-2\fboxrule\relax}{%
\textbf{Finding 4.} \textit{API-level knowledge transfer improves compilation success, but overlooks strict on-chip capacity limits, leading to inevitable runtime crashes during kernel execution.}
}}

\vspace{4pt}

To validate this, we design a category-shot prompting strategy using LeakyReLU as the in-context example. Its API footprint heavily overlaps with the 15 target activation kernels, yet its exclusion from the target set forces the LLM to genuinely generalize rather than trivially copy. As shown in Figure \ref{fig:prompt_strategy}, this strategy boosts Comp@1 by 33–47\% over the baseline. However, Pass@1 catastrophically drops to 0\%. Analysis reveals a single root cause for these runtime crashes: Unified Buffer (UB) overflow.

The UB is a strictly limited 256KB on-chip workspace. The LLMs blindly replicate LeakyReLU's tiling parameters, safely allocating 192KB for two buffers. However, kernels like Softmax require extra intermediate buffers, driving allocation to 384KB and immediately violating the capacity limit.

By explicitly providing the 256KB constraint and a tiling computation formula (category-shot+UB-hint), Pass@1 is restored to 20.0\% (DeepSeek-V4 Pro) and 33.3\% (GLM-4.5) (Figure \ref{fig:prompt_strategy}). With functional correctness validated, we now turn our focus to the next critical objective: execution performance.

\begin{table}[t]
\centering
\setlength\abovecaptionskip{0.02\baselineskip}
\setlength\belowcaptionskip{-0.5\baselineskip}
\caption{Results on RTX4090D and 910B2. ``--'' = the LLM never generates a correct result.}
\label{tab:perf_matmul}
\scriptsize
\begin{tabular}{p{1cm}ccccc}
\hline
\textbf{Op.}  & \textbf{CUDA} & \textbf{PyT. (NPU)} & \textbf{Asc. C} & \textbf{Norm.} & \textbf{Slow.} \\
 & \textbf{(ms)} & \textbf{(ms)} & \textbf{(ms)} & \textbf{vs CUDA} & \textbf{vs PyT.} \\
\hline
diagonal  & 0.2 & 1.7 & 37.4 & 295.6$\times$ & 22.0$\times$ \\
trans\_A  & 4.6 & 1.7 & 37.4 & 10.2$\times$ & 22.5$\times$ \\
lower\_tri  & 1.1 & 1.7 & 37.5 & 45.2$\times$ & 21.7$\times$ \\
batched  & 10.5 & 3.7 & -- & -- & -- \\
square  & 5.1 & 1.7 & 37.4 & 9.2$\times$ & 22.7$\times$ \\
irregular  & 11.1 & 4.3 & -- & -- & -- \\
4D  & 18.8 & 5.7 & -- &  -- & -- \\
tall\_skinny  & 9.0 & 3.2 & 33.4 & 4.7$\times$ & 10.4$\times$ \\
3D  & 1.7 & 0.6 & -- & -- & -- \\
large\_K  & 27.0 & 2.4 & -- & -- & -- \\
small\_K  & 10.3 & 3.4 & 38.0 & 4.7$\times$ & 11.2$\times$ \\
standard  & 5.5 & 1.6 & 37.4 & 8.6$\times$ & 22.8$\times$ \\
trans\_B  & 4.5 & 1.7 & 37.8 & 10.7$\times$ & 22.6$\times$ \\
mat\_vec  & 9.0 & 36.2 & -- & -- & -- \\
trans\_both  & 5.3 & 1.7 & 37.6 & 8.9$\times$ & 22.7$\times$ \\
symmetric  & 4.6 & 1.7 & 37.4 & 10.4$\times$ & 22.7$\times$ \\
upper\_tri  & 1.1 & 1.7 & 37.6 & 44.4$\times$ & 21.9$\times$ \\
\hline
\textit{Avg.} &  &  & & 41.1$\times$ & 20.3$\times$ \\
\hline
\end{tabular}
\vspace{-15pt}
\end{table}

\noindent\fcolorbox{black}{gray!8}{\parbox{\dimexpr\columnwidth-2\fboxsep-2\fboxrule\relax}{%
\textbf{Finding 5.} \textit{Functionally correct NPU kernels suffer severe performance degradation, as LLMs fail to leverage architecture-specific acceleration features without explicit optimization priors.}
}}

Even when LLMs generate functionally correct Ascend C kernels, their execution performance is often abysmal. We evaluate 17 matmul kernels from MultiKernelBench using category-shot+UB-hint on a 910B2 versus an RTX 4090D generated by GLM-4.5. As shown in Table \ref{tab:perf_matmul}, while all 17 CUDA kernels execute correctly, only 11 Ascend C kernels pass. Critically, these 11 kernels run average 20.3$\times$ slower than the implementation via PyTorch on 910B2. After normalizing the theoretical compute capacity between the 910B2 and 4090D, the Ascend kernels exhibit a average 41.1$\times$ slowdown relative to their CUDA counterparts, proving the bottleneck lies in deficient code optimization rather than hardware capability. Code inspection reveals two naive anti-patterns driving this degradation: restricting execution to a single AI Core (\texttt{SetBlockDim(1)}) and failing to enable high-throughput HF32 mode (\texttt{matmulObj.SetHF32(true)}).

Based on this analysis, we simply append a one-line HF32 hint to the prompt for three representative kernels. As Table \ref{tab:hf32_hint} shows, this minimal optimization hint reduces latency by 27.1\% on average, confirming that explicit architectural priors are crucial for high-performance generation.

\vspace{-5pt}

\subsection{Strawman Idea}

\begin{table}[t]
\centering
\caption{HF32 performance hint effectiveness on Ascend 910B2.}
\label{tab:hf32_hint}
\scriptsize
\begin{tabular}{p{2.0cm}cccc}
\hline
\textbf{Op.} & \textbf{Ascend C} & \textbf{Optimized} & \textbf{$\Delta$} & \textbf{Improv.} \\
 & \textbf{(ms)} & \textbf{(ms)} & \textbf{(ms)} & \\
\hline
symmetric & 37.4 & 27.1 & $-$10.3 & 27.5\% \\
square & 37.4 & 26.9 & $-$10.6 & 28.2\% \\
standard & 37.4 & 27.8 & $-$9.6 & 25.7\% \\
\hline
\textit{Avg.} & \textit{37.4} & \textit{27.3} & \textit{$-$10.2} & \textit{27.1\%} \\
\hline
\end{tabular}
\vspace{-20pt}
\end{table}

Based on our empirical progression (\S \ref{motivation}), we categorize the missing priors causing LLM generation failures into three pillars: API-specific syntax, strictly bounded capacity constraints, and architecture-specific optimization features. We collectively define these as hardware-aware knowledge. The findings reveal a clear conclusion: explicitly harnessing this unified, domain-specific knowledge is the key to rescuing LLMs from NPU kernel generation failures and unlocking high performance. Based on this conclusion, a natural strawman approach is to dynamically harness hardware-aware knowledge based on empirical execution feedback during kernel generation. However, operationalizing this immediately encounters three formidable challenges. 

\textbf{(1) Heterogeneous Structuring}, as demonstrated by the \textit{category-shot} failures (\S \ref{subsec:NPUfinding}), blindly reusing raw code snippets triggers runtime crashes because they lack explicit rationales and hardware boundaries. Knowledge must span diverse modalities (prose, pseudo-code, templates) without exhausting the LLM's token budget through naive concatenation or destroying syntactic precision through homogenization. \textbf{(2) Hardware-Aware Retrieval}, standard embedding-based retrieval relying on textual or functional similarity is fundamentally inadequate. Operators sharing identical mathematical functions often diverge completely in their underlying hardware bottlenecks, necessitating a retrieval mechanism that jointly evaluates exact syntactic API matches and deep, hardware-aligned semantic constraints. \textbf{(3) Context-Aware Maintenance}, the knowledge base inevitably accumulates erroneous entries and redundant patterns. Because the validity of any optimization rule is strictly bound to its specific execution context, maintaining a compact, high-purity corpus demands an automated distillation process rigorously grounded in actual compilation and execution outcomes.

\section{Hawk Overview}

To tackle the intricacies of NPU kernel generation, we introduce Hawk, a training-free framework built upon a foundational code agent that seamlessly harnesses hardware-aware knowledge into existing development pipelines. Specifically, we implement Hawk as an advanced extension of CANNBot~\cite{cannbot}, an existing baseline system designed to assist code agents in NPU operator generation. While CANNBot provides standard orchestration and official API documentation, it lacks the implicit hardware constraints necessary for correct NPU execution—a critical performance gap that Hawk's hardware-aware knowledge directly addresses. As shown in Figure \ref{MemKernel_overview}, Hawk operates in two primary phases:

\begin{figure}[t]
    \centering
    \setlength\abovecaptionskip{0.02\baselineskip}
    \setlength\belowcaptionskip{-0.5\baselineskip}
	\includegraphics[width=\linewidth]{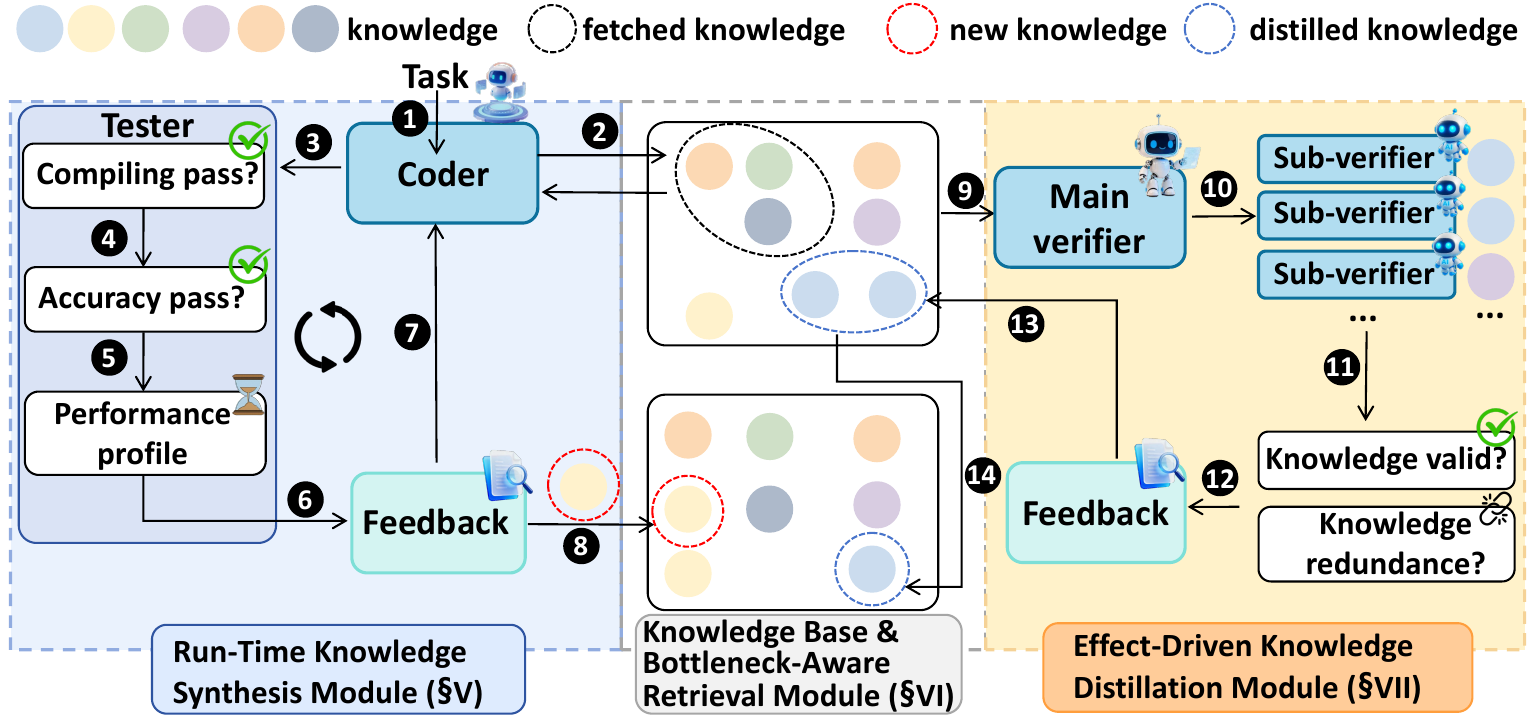}
	\caption{Overview of Hawk.}
    \label{MemKernel_overview}
    \vspace{-15pt}
\end{figure}

Phase 1: Knowledge-Guided Generation and Synthesis (Steps 1–8). The foundation of Hawk is the Run-Time Knowledge Synthesis Module, which structures hardware expertise into a hierarchical representation (keywords, rationales, and code templates) and dynamically formulates new insights. Upon receiving a kernel generation task, the code agent leverages the Bottleneck-Aware Retrieval Module to query this structured knowledge base, fetching hardware-aligned patterns rather than superficial API matches. Equipped with this precise knowledge, the agent generates the kernel, executes it, and analyzes the empirical feedback to iteratively refine the code. The synthesis module abstracts the newly resolved hardware constraints during the kernel generation process, enriching the knowledge base for future tasks.

Phase 2: Knowledge Distillation (Steps 9–14). To prevent the knowledge base from degrading as the system scales, Hawk employs the Effect-Driven Knowledge Distillation Module. A primary verification agent orchestrates multiple sub-agents to validate knowledge entries in parallel across diverse operator contexts. Based on these collective empirical reports, the system automatically purges erroneous entries that trigger crashes and merges redundant patterns that monopolize the LLM context. This continuous distillation ensures the knowledge base remains highly accurate.

\vspace{-5pt}

\section{Run-Time Knowledge Synthesis Module}

To systematically capture high-value hardware priors without polluting the LLM's context window, Hawk replaces naive output logging with a structured schema and rigorous trigger mechanisms.

\textbf{Triple-Part Executable Knowledge Representation.} As highlighted in Challenge 1, simply concatenating raw code snippets exhausts token budgets and lacks hardware rationale, while homogenizing formats destroys syntactic precision. To reconcile semantic expressiveness with syntactic fidelity, Hawk structures each knowledge entry into a tripartite hierarchy: 
\textit{(1) Metadata Layer (Indexing):} Captures operational context (e.g., \texttt{operator\_type, data\_type}, \texttt{cann\_version}) and hardware bottleneck tags (e.g., bandwidth, tiling).
\textit{(2) Rationale Layer (Semantics): }Provides natural language justifications backed by profiling data, explicitly explaining \textit{why} and \textit{when} an optimization applies (e.g., specific hardware boundaries). 
\textit{(3) Artifact Layer (Reference Code):} Stores the exact, executable Ascend C snippet. 
This design ensures the LLM learns hardware constraints besides mere syntax.

\label{Triggering-Conditions}
\textbf{Rigorous Triggering Conditions.} To guarantee that the agent captures only high-value micro-architectural insights rather than trivial code adjustments, knowledge synthesis is triggered \textit{if and only if} an iteration during kernel generation satisfies one of four strict criteria: 
\textit{(1) Iterative Difficulty:} Resolving non-trivial bugs requiring $\geq$2 refinement rounds. 
\textit{(2) Ground Truth Correction:} Uncovering mismatches between official API documentation and actual hardware behavior. 
\textit{(3) Quantifiable Performance Gain:} Applying optimizations that yield measurable speedups beyond a predefined threshold. 
\textit{(4) Design Deviation:} Adopting mathematically equivalent but structurally superior derivations that outperform standard API calls.

\textbf{Cold-Start Data Pre-Precipitation.} To bootstrap Hawk from a zero-knowledge state, we construct an initial knowledge base via an offline, three-stage pipeline mining human-expert kernels: \textit{Stage 1: Oracle Construction.} We automatically generate 20 comprehensive test cases per kernel, covering extreme tensor shapes and boundaries for rigorous functional verification. \textit{Stage 2: AI-Driven Autonomous Tuning.} The agent generates an initial kernel from natural language and enters an iterative refinement loop. Guided by hardware profiling (e.g., using MTE2-to-VEC ratios to identify memory vs. compute bottlenecks), it autonomously optimizes the code to pass all tests while maximizing performance. \textit{Stage 3: Expert-Guided Differential Analysis.} Upon hitting a performance plateau, the agent compares its code against a human-expert reference. By analyzing structural gaps, it explicitly optimizes its kernel toward the expert's performance ceiling. Finally, all verified optimizations from Stages 2 and 3 are systematically distilled, equipping Hawk with a hardware-aware corpus.


\section{Bottleneck-Aware Retrieval Module}

To address the retrieval challenge (C2), Hawk operationalizes a \textit{2D-Retrieval mechanism} that fuses syntactic and semantic dimensions, followed by a feedback-driven iterative loop to prevent context exhaustion.


\subsection{2D-Retrieval Mechanism}

 Relying solely on syntactic matching fetches generic API templates but fails to capture the hardware bottlenecks, leading to runtime inefficiencies. Conversely, pure semantic similarity often retrieves conceptually analogous but syntactically incompatible APIs, triggering immediate compilation failures. To guarantee both syntactic compilability and hardware compliance, a retrieval mechanism must evaluate these dual constraints simultaneously. Therefore, we decompose knowledge relevance into two complementary dimensions and fuse them into a unified scoring function $\mathcal{S}(K_i, Q)$ for a knowledge entry $K_i$ and query $Q$:

\textbf{(1) Syntactic Exactness (syn):} Ascend C development is strictly bound to specific hardware identifiers, API signatures, and error codes (e.g., \texttt{ACL\_ERROR\_xxx}). We employ the BM25 algorithm~\cite{askari2023injecting} to capture these exact keyword matches. This ensures that if the agent encounters a specific API crash, the retriever prioritizes knowledge tagged with that exact identifier, providing a syntactic safety net.

\textbf{(2) Hardware-Aligned Semantic Similarity (sem):} Relying solely on lexical matching is brittle because agent feedback is often high-level natural language, which may not share keywords with the actual solution. We utilize dense embeddings~\cite{zhang2025qwen3} to encode query and knowledge entry. Cosine similarity between these vectors captures the underlying hardware bottlenecks even when vocabularies diverge.

A critical challenge in fusion is that score distributions differ vastly (BM25 is unbounded, while cosine similarity is bounded in $[-1, 1]$). Simple weighted summation requires tedious, brittle hyperparameter tuning. To solve this, we adopt Reciprocal Rank Fusion (RRF)~\cite{cormack2009reciprocal}, operating on ranks rather than raw scores:

\begin{equation}
    \mathcal{S}(K_i, Q) = \sum_{r \in \{\text{syn}, \text{sem}\}} \frac{1}{k + \text{rank}_r(K_i)}
\end{equation}
Here, $\text{rank}_r(K_i)$ denotes the rank position of knowledge $K_i$ within the dimension $r$. Following existing works~\cite{rackauckas2024rag, samuel2025mmmorrf}, we set the smoothing constant $k=60$ balances top-ranked and lower-ranked candidates. RRF normalizes the two dimensions into reciprocal ranks, eliminating manual weight tuning and robustly fetching hardware-aware patterns.

\vspace{-5pt}

\subsection{Feedback-Driven Iterative Retrieval}
After computing $\mathcal{S}(K_i, Q)$, the agent consumes the knowledge base via an adaptive loop. In iteration $t$, the agent requests the Top-$K$ candidates. If applying these candidates fails to resolve the hardware bottleneck, it indicates the original query $Q_t$ was misaligned with the root cause. Under this case, the agent triggers a \textit{Feedback-Driven Query Reformulation} phase. It synthesizes a refined query $Q_{t+1}$ using execution feedback via two structured mechanisms: (1) \textit{Diagnostic Semantic Shift:} Analyzing the error trace to shift search focus. For example, if a retrieved memory-tiling strategy causes an out-of-order fault, the agent shifts the semantic focus from ``memory allocation'' to ``pipeline synchronization primitives.'' (2) \textit{Contrastive Exclusion:} Encoding the failed strategies as negative constraints in $Q_{t+1}$ to steer the retriever away from ineffective solution spaces. The system then recalculates $\mathcal{S}(K_i, Q_{t+1})$ for a fresh Top-$K$ batch, iterating until the bottleneck is resolved.

\vspace{-5pt}

\section{Effect-Driven Knowledge Distillation}

To guarantee the long-term quality and contextual efficiency of the knowledge base (Addressing C3), Hawk incorporates an effect-driven distillation process. Unlike traditional static AST-based cleaning, Hawk leverages empirical execution feedback and LLM-driven semantic arbitration to eliminate misdirection and halt context window exhaustion.

\textbf{Empirical Verification and Boundary Refinement.} During the generation loop, if an applied knowledge-infused code snippet triggers a compilation or runtime failure, a validation sub-agent is dispatched to cross-examine the anomaly. Based on the empirical error trace, it applies a targeted resolution: if the knowledge is fundamentally flawed, it is permanently purged to prevent future misdirection. Conversely, if it is valid but misapplied to the current operator, the sub-agent explicitly appends usage boundaries to its metadata layer (e.g., restricting a tiling strategy to specific tensor sizes). This continuous cycle guarantees that every insight remains rigorously bound to its applicable operational context.

\begin{figure*}[t]
    \centering
    \setlength\abovecaptionskip{0.02\baselineskip}
    \setlength\belowcaptionskip{-0.5\baselineskip}
	\includegraphics[width=\linewidth]{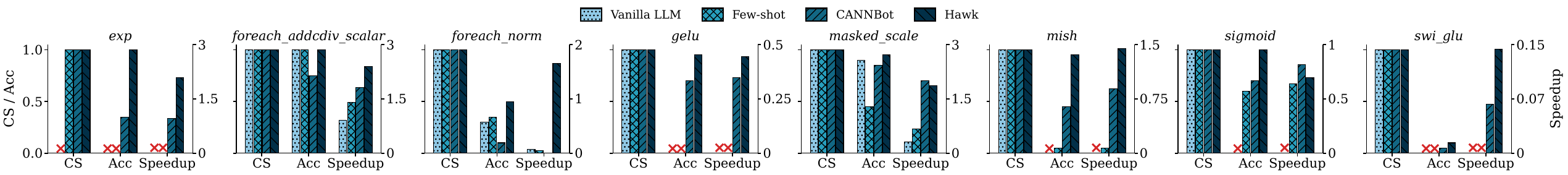}
	\caption{The overall performance of L1 kernels with GLM-5.1.}
    \label{exp:exp_overview}
    \vspace{-20pt}
\end{figure*}

\textbf{Conflict Arbitration.} As the system scales, contradictory strategies inevitably emerge. Periodically, an Arbitrator Agent samples knowledge pairs $(k_a, k_b)$ sharing overlapping tags. If $k_a$ and $k_b$ suggest logically conflicting optimizations—for instance, one mandates a \texttt{PipeBarrier} for conservative synchronization while the other omits it to maximize pipeline overlap—the Arbitrator actively investigates their empirical execution outcomes. Based on this investigation, the Arbitrator enforces one of two resolutions: (1) \textit{Pruning:} If one entry is strictly superior or contains factual errors, the inferior entry is discarded. (2) \textit{Contextual Merging:} If both are conditionally correct but practically conflict due to overlapping scopes, the Arbitrator refines their boundary conditions to ensure mutual exclusivity, transforming a conflict into a comprehensive, conditional rule.

\textbf{Semantic Consolidation.} To prevent retrieval noise from near-duplicate entries, a Merger Agent periodically consolidates functionally overlapping knowledge. Instead of simple text concatenation, the Merger executes a three-step semantic fusion: (1) \textit{Aligning Common Ground:} Extracting shared optimization baselines without introducing contradictions; (2) \textit{Integrating Divergences:} Encoding varying operational trade-offs (e.g., precision formats) into unified conditional branches; and (3) \textit{Synthesizing Rationale:} Generating a singular, non-redundant explanation to improve LLM decision traceability. The resulting consolidated entry supersedes the originals, effectively reducing database bloat and providing the primary agent with comprehensive guidance without being overwhelmed by repetitive context.

\vspace{-4pt}

\section{Experiment}

\subsection{Experiment Setup}

\textbf{Implementation.} We implement Hawk as custom agent skills within CANNBot. CANNBot provides the foundational architecture: (1) role-based workflow orchestration (designer, developer, evaluator), (2) standardized agent skills for CANN tools, and (3) \textit{asc-devkit} official documentation for hardware information and templates. To prevent data leakage, we initialize Hawk's knowledge base using 158 distilled entries derived from \textit{ops-nn}~\cite{opsnn} L1 kernels, which are strictly disjoint from our test set categories. 

\textbf{Benchmark and Metrics.} We evaluate Hawk on L1 and L2 kernels from CANNBench, a benchmark providing 20 test cases per Ascend C operator. We evaluate performance using four metrics: \textbf{(1) Compilation Success (CS)} (binary compilation status); \textbf{(2) Correctness (Acc)} (percentage of passed test cases); \textbf{(3) Speedup} ($L_{std}/L_{opt}$, where $L_{std}$ and $L_{opt}$ are the execution latencies of the standard implementation provided by CANNBench and the generated kernel implementation, respectively); and \textbf{(4) Score} (a comprehensive official rating out of 100):
\begin{equation}
    Score = 20 \times CS + 30 \times Acc + 50 \times PR
\end{equation}
where $PR$ (Performance Rate) evaluates latency against the hardware's theoretical lower bound ($L_{hw}$):
\begin{equation}
    PR = \frac{L_{std} - L_{hw}}{(L_{opt} - L_{hw}) + (L_{std} - L_{hw})}
\end{equation}
\textit{(Note: $PR$ and Speedup are averaged across the 20 test cases. Score can exceed 100 if $L_{opt} < L_{hw}$. The score weight for each metric follows the official document.)}

\textbf{Baselines.} Using Claude Code as the universal base agent, we compare Hawk against three baselines: \textbf{(1) Vanilla LLM}~\cite{zeng2026glm,xu2026deepseek} (bare-bones Claude Code), \textbf{(2) Few-Shot}~\cite{xu2024does} (Claude Code assisted by standard code snippets), and \textbf{(3) CANNBot}~\cite{cannbot} (the base plugin utilizing only standard tools and official manuals).

\textbf{Experiment Details.} All experiments are conducted on a machine with 8 910B2 NPUs running CANN version 8.5.1. For the Few-shot baseline, the prompt is dynamically augmented with a reference implementation retrieved based on the functional semantic similarity of operator descriptions encoded by the Qwen3-embedding-8B model~\cite{zhang2025qwen3}. To strictly prevent data leakage, the target operator is explicitly excluded from the retrieval corpus. For Hawk, the retrieval module fetches $K$=6 knowledge entries per iteration—a parameter empirically determined in our ablation study to optimally balance hardware-aware guidance with token economy. To ensure a fair comparison, we uniformly cap the maximum agent iterations at 30 across all evaluated methods. We use GLM-5.1 as the LLM backbone for Claude Code.

\begin{figure*}[t]
    \centering
    \setlength\abovecaptionskip{0.02\baselineskip}
    \setlength\belowcaptionskip{-0.5\baselineskip}
	\includegraphics[width=\linewidth]{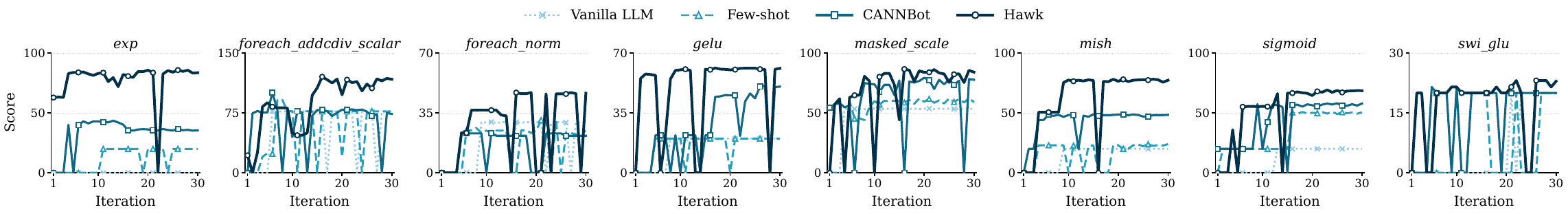}
	\caption{The overview of the Score trajectory throughout the L1 kernel generation iterations with GLM-5.1.}
    \label{exp:iteration}
    \vspace{-20pt}
\end{figure*}

To rigorously evaluate our framework, we formulate three core Research Questions (RQs) that assess Hawk's overall effectiveness, the contribution of its internal mechanisms, and its scalability:

\begin{itemize}
    \item \textbf{RQ1:} How does Hawk compare to baseline methods in generating functionally correct and high-performance NPU kernels?
    \item \textbf{RQ2:} How do the individual components of Hawk contribute to its generation capabilities?
    \item \textbf{RQ3:} How well does Hawk scale to complex kernels and generalize across weaker LLM backbones?
\end{itemize}

\subsection{RQ1: Overall Performance}

\textbf{Overall Correctness and Performance.} Figure~\ref{exp:exp_overview} compares Hawk against three baselines across eight NPU operators. While Few-shot achieves Compilation Success on multiple operators, it fundamentally fails at execution (e.g., 0\% Acc on \textit{exp} and \textit{gelu}) by merely mimicking surface-level syntax without semantic understanding. CANNBot improves correctness on simpler tasks but plummets on complex operators (e.g., 10\% Acc on \textit{foreach\_norm}). Relying solely on official manuals, it misses implicit hardware constraints (e.g., Unified Buffer limits), resulting in dismal execution efficiency (e.g., 0.007$\times$ Speedup) even when the generated code is functionally correct. 

In contrast, Hawk seamlessly bridges syntactic validity and hardware efficiency. By decoupling executable templates from optimization rationales, it systematically guides the agent to respect underlying hardware constraints. Consequently, Hawk achieves robust Accuracy (100\% on \textit{exp} and \textit{foreach\_addcdiv\_scalar}) and delivers exceptional Speedups (averaging 2.10$\times$ and 2.39$\times$ over 20 test cases, respectively). Although Hawk performs lower Speedup against CANNBot on \textit{masked\_scale}, \textit{sigmoid} operators, it achieves higher Acc across all test cases, demonstrating that Hawk prioritizes functional correctness and numerical stability over aggressive but potentially unsafe optimizations.

\begin{figure}[t]
    \centering
    \setlength\abovecaptionskip{0.02\baselineskip}
    \setlength\belowcaptionskip{-0.5\baselineskip}
    \begin{minipage}[t]{0.48\columnwidth}
      \centering
      \includegraphics[width=\linewidth]{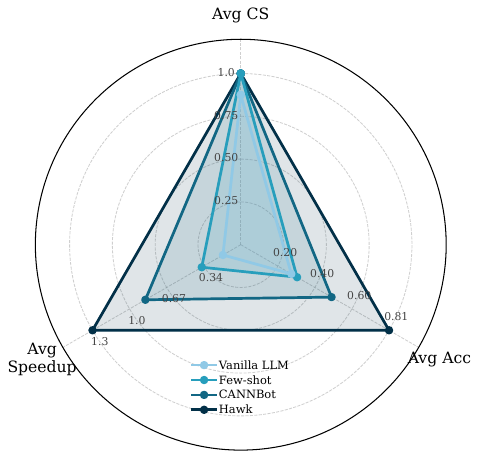}
      \caption{Normalized capability.}
      \label{exp:radar}
    \end{minipage}\hfill
    \begin{minipage}[t]{0.48\columnwidth}
      \centering
      \includegraphics[width=\linewidth]{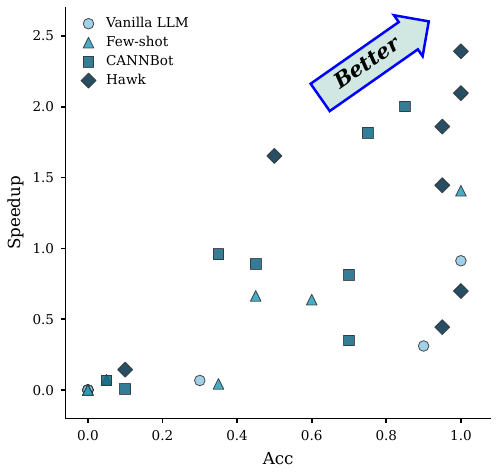}
      \caption{Acc vs Speedup.}
      \label{exp:Acc_vs_Speedup}
    \end{minipage}
    \vspace{-20pt}
  \end{figure}

\textbf{Comprehensive Evaluation.} Figure \ref{exp:radar} highlights Hawk's holistic superiority, demonstrating that it achieves a higher average Acc (0.80) and a faster average execution Speedup (1.34$\times$) across the evaluated L1 operators, significantly outperforming the strongest baseline, CANNBot (0.49 Acc and 0.86$\times$ Speedup). To further disentangle the relationship between correctness and speed, Figure \ref{exp:Acc_vs_Speedup} explicitly maps the accuracy-speedup trade-off. Hawk achieves superior execution performance built upon a foundation of high accuracy.

\textbf{Iterative Trajectory Analysis.} As shown in Figure \ref{exp:iteration}, Hawk can achieve high score quicklier than other baselines. Taking \textit{exp} as an example, the Vanilla LLM scores 0 throughout, failing compilation entirely. Few-shot slowly reaches compilable syntax at iteration 11 (score 20) but never achieves functional correctness. CANNBot shows faster initial progress, reaching correctness by iteration 4 (score 40.11), but permanently plateaus as it lacks debugging heuristics to escape local optima. Conversely, Hawk achieves compilation and correctness on its very first attempt (initial score 62.82). In contrast, by dynamically translating runtime feedback into targeted semantic queries, Hawk swiftly discovers hardware optimizations, rapidly accelerating to a peak score of 83.82 by iteration 5 and maintaining this highly optimized state.

\textit{RQ1 summary: } Hawk decisively bridges the gap between syntactic validity and hardware efficiency. It achieves up to 100\% Accuracy on complex operators and delivers Speedups averaging up to 2.39$\times$ across 20 test cases, overcoming the near-zero efficiency of baselines.

\subsection{RQ2: Ablation Study}

We conduct a thorough ablation study (Table~\ref{Ablation_study}) to isolate the impact of Hawk's core modules.

  \begin{figure*}[t]
    \centering
    \setlength\abovecaptionskip{0.02\baselineskip}
    \setlength\belowcaptionskip{-0.5\baselineskip}
	\includegraphics[width=\linewidth]{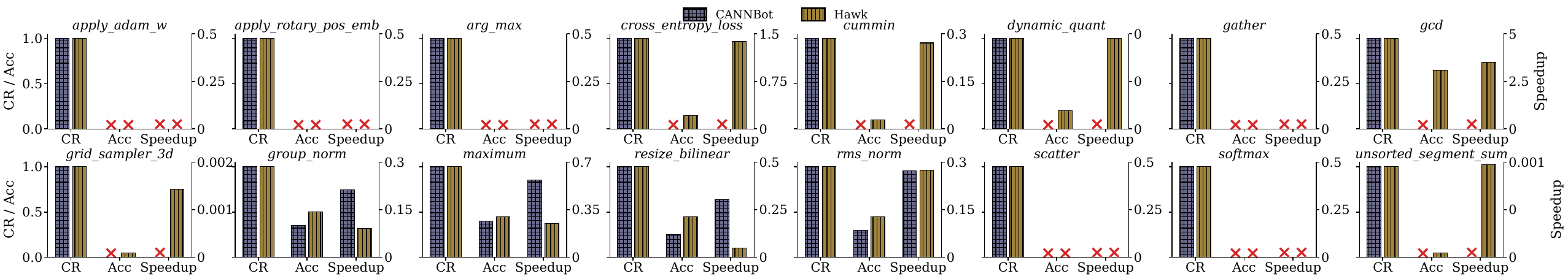}
	\caption{The overall performance of L2 kernels with GLM-5.1.}
    \label{exp:exp_l2}
    \vspace{-10pt}
\end{figure*}

  \begin{figure*}[t]
    \centering
    \setlength\abovecaptionskip{0.02\baselineskip}
    \setlength\belowcaptionskip{-0.5\baselineskip}
    \begin{minipage}[c]{0.35\textwidth}
      \centering                     
      \includegraphics[width=\linewidth]{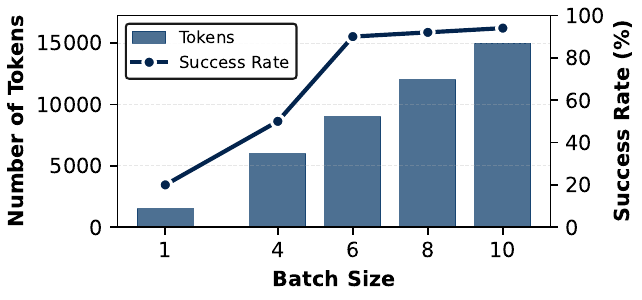}\\[2pt]
      \caption{Batch size analysis.}
      \label{fig:topk}
    \end{minipage}\hfill                                        
    \begin{minipage}[c]{0.61\textwidth}
      \centering
      \includegraphics[width=\linewidth]{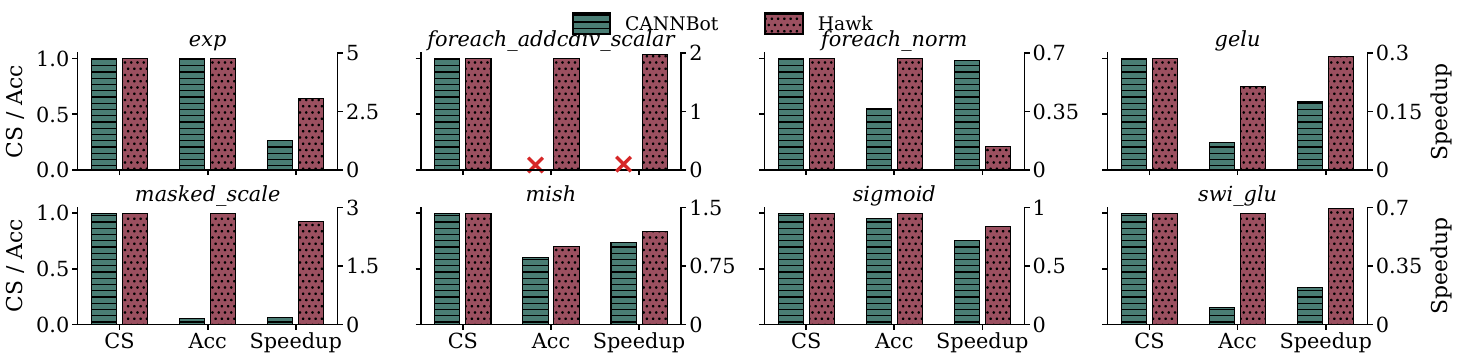}\\[2pt]
      \caption{The L1 kernel performance with DeepSeek-V4-Flash.}
      \label{fig:deepseek}
    \end{minipage}
    \vspace{-20pt}
  \end{figure*}    

\textbf{Impact of Knowledge Structuring.} When we downgrade Hawk's triple-part hierarchical structure to unstructured plain-text descriptions, Acc drops catastrophically from 0.84 to 0.10, and Speedup collapses to 0.09. This profound degradation confirms that LLMs cannot reliably extract operational boundaries from dense prose. 

\begin{table}[t]
\caption{Ablation study of Hawk.}
\centering
\setlength\abovecaptionskip{0.02\baselineskip}
\setlength\belowcaptionskip{-0.5\baselineskip}
\label{Ablation_study}
\begin{tabular}{@{}lccc@{}}
\toprule
\multicolumn{1}{c}{\textbf{Variant}} & \multicolumn{1}{c}{\textbf{CR}} & \multicolumn{1}{c}{\textbf{Acc}} & \multicolumn{1}{c}{\textbf{Speedup}} \\ \midrule
Hawk-Full                            & 1.00                              & 0.84                             & 1.00$\times$                          \\
w/o Structured Representation        & 0.88                           & 0.10                              & 0.09$\times$                            \\
w/o 2D-Retrieval                     & 1.00                               & 0.45                             & 0.41$\times$                            \\
w/o Distillation                     & 1.00                               & 0.28                             & 0.25$\times$                         \\ \bottomrule
\end{tabular}
\vspace{-20pt}
\end{table}

\textbf{Impact of 2D-Retrieval Mechanism.} Replacing our 2D-Retrieval with standard plain-text semantic matching causes Acc to decline to 0.45 and Speedup to 0.41. Standard semantic search retrieves knowledge based on symptom descriptions rather than the structural root cause. The 2D-Retrieval mechanism's fusion of syntactic constraints and hardware-aligned semantics ensures that the retrieved knowledge is not just conceptually similar, but functionally actionable for the bottleneck.

\textbf{The Necessity of Distillation.} To evaluate the distillation module, we reverted the knowledge base to its raw, pre-distilled state (expanding it slightly from 158 to 165 entries). Astoundingly, the addition of merely 7 unverified entries (a 4.4\% size increase) caused Acc to plummet from 0.84 to 0.28, and Speedup to drop from 1.00 to 0.25. Even a tiny fraction of conflicting or contextually mismatched knowledge can completely derail the LLM's reasoning chain. Distillation is therefore not merely a storage optimization, but a vital defense mechanism against logical contradictions.

\textbf{Top-$K$ Retrieval Efficiency.} Figure~\ref{fig:topk}  reveals a clear trade-off between first-pass success rate and token budget. Small batches ($K < 6$) lead to low success rates ($<55\%$) and excessive, costly query reformulations. Conversely, $K$=6 yields marginal gains (+2\% over $K$=6) while significantly inflating the prompt context, increasing the risk of distraction. We identify $K$=6 as the optimal balance, achieving a 90\% first-pass success rate with high token economy.

\textit{RQ2 summary: } Every core module is indispensable. Disabling hierarchical structuring or 2D-Retrieval collapses Acc from 0.84 to 0.10 and 0.45, respectively. Without distillation, minor unverified noise (4.4\%) poisons the context, dropping Acc to 0.28. Conversely, the retrieval batch size $K$=6 is a relatively suitable choice.

\vspace{-5pt}

\subsection{RQ3: Scalability of Hawk}

\textbf{Scalability from L1 to L2 Kernels.} A robust knowledge framework should allow foundational principles to transfer to higher-order problems. We evaluated Hawk on 16 complex L2 operators using the knowledge base exclusively mined from simpler L1 operators (Figure~\ref{exp:exp_l2}). Both CANNBot and Hawk achieved 100\% compilation, but CANNBot yielded 0\% Acc on 12 out of 16 operators. In contrast, Hawk delivered substantial functional improvements on 8 operators (e.g., 65\% Acc on \textit{gcd}) and achieved speedups up to 3.5$\times$. This disparity demonstrates that despite differing mathematical logic, L2 kernels share overlapping hardware constraints with L1. Hawk effectively transfers prior L1 knowledge to resolve these shared bottlenecks.

\textbf{Generalizability to Weaker LLM Backbones.} To determine if Hawk is overfitted to highly capable models, we evaluated its transferability on a weaker backbone, DeepSeek-V4-Flash. Figure \ref{fig:deepseek} indicates that, compared to the CANNBot baseline, Hawk significantly elevated the average Acc from 44.4\% to 93.1\% and improved the average Speedup from 0.53$\times$ to 1.35$\times$. This demonstrates Hawk can generalize to weaker LLMs.

\textit{RQ3 summary: } Hawk exhibits exceptional scalability, transferring hardware priors from L1 to complex L2 kernels to boost Accuracy from 0\% to 65\% alongside up to 3.5$\times$ Speedups. It demonstrates generalizability, elevating a weaker DeepSeek backbone's Accuracy (15\%$\rightarrow$60\%) and average Speedup (0.2$\times$$\rightarrow$0.8$\times$).


\section{Threats to Validity}


\textbf{Internal Validity.} The primary internal threats are data leakage and human selection bias. Specifically, the coding agent might trivially bypass the generation task by directly copying ground-truth implementations exposed in the knowledge base. Additionally, researchers might intentionally cherry-pick optimization rules explicitly tailored for the test set. To eliminate the severe threat of data leakage or human selection bias, Hawk's knowledge base was populated exclusively using \textit{ops-nn} kernels that are strictly disjoint from the CANNBench test set. Furthermore, the entire knowledge extraction pipeline was automated via predefined empirical triggering conditions (\S\ref{Triggering-Conditions}), ensuring the distillation process remained entirely objective.

\textbf{External Validity.} External threats primarily concern the generalizability of our framework across different NPU hardware generations, LLM backbones, and operator complexities. Although our primary evaluation was instantiated on a specific Ascend NPU version, Hawk's core methodology—hierarchical knowledge structuring and bottleneck-aware retrieval—is rooted in the unified Da Vinci architecture. Consequently, it can seamlessly generalize to other Ascend NPU series (e.g., Ascend 310, 950) that share the CANN toolchain and similar memory hierarchies. Moreover, our supplementary evaluation on DeepSeek-V4-Flash definitively confirms that Hawk's effectiveness is not isolated to a specific model. Finally, while our test suite covers comprehensive L1 and L2 kernels, future work will explore scaling this knowledge-driven paradigm to highly complex fused operators on advanced NPU clusters.

\vspace{-5pt}

\section{Conclusion}

This paper presents Hawk, a training-free framework that harnesses hardware-aware knowledge for high-performance NPU kernel generation. By proposing hierarchical knowledge synthesis, bottleneck-aware retrieval, and effect-driven distillation, Hawk raises the accuracy rate from 49.4\% to 80\%, while achieving up to 2.2$\times$ speedup against the SOTA works.

\bibliographystyle{IEEEtran}
\bibliography{sample}

@String{Computing = "Computing" }

@String{Computer = "{IEEE} Computer" }

@String{Springer = "Springer-Verlag" }

@article{cao2026ascendkernelgen,
  title={AscendKernelGen: A Systematic Study of LLM-Based Kernel Generation for Neural Processing Units},
  author={Cao, Xinzi and Zhai, Jianyang and Li, Pengfei and Hu, Zhiheng and Yan, Cen and Mu, Bingxu and Fang, Guanghuan and She, Bin and Li, Jiayu and Su, Yihan and others},
  journal={arXiv preprint arXiv:2601.07160},
  year={2026}
}

@article{zheng2026towards,
  title={Towards Cold-Start Drafting and Continual Refining: A Value-Driven Memory Approach with Application to NPU Kernel Synthesis},
  author={Zheng, Yujie and Li, Zhuo and Zhang, Shengtao and Wang, Hanjing and Sheng, Junjie and Wang, Jiaqian and Yan, Junchi and Zhang, Weinan and Wen, Ying and Tang, Bo and others},
  journal={arXiv preprint arXiv:2603.10846},
  year={2026}
}

@article{wen2026ascendcraft,
  title={AscendCraft: Automatic Ascend NPU Kernel Generation via DSL-Guided Transcompilation},
  author={Wen, Zhongzhen and Shao, Shudi and Li, Zhong and Ge, Yu and Xu, Tongtong and Lin, Yuanyi and Zhang, Tian},
  journal={arXiv preprint arXiv:2601.22760},
  year={2026}
}

@inproceedings{zhao2021akg,
  title={AKG: automatic kernel generation for neural processing units using polyhedral transformations},
  author={Zhao, Jie and Li, Bojie and Nie, Wang and Geng, Zhen and Zhang, Renwei and Gao, Xiong and Cheng, Bin and Wu, Chen and Cheng, Yun and Li, Zheng and others},
  booktitle={Proceedings of the 42nd ACM SIGPLAN International Conference on Programming Language Design and Implementation},
  pages={1233--1248},
  year={2021}
}

@software{kernelllm2025,
    title={KernelLLM: Making Kernel Development More Accessible},
    author={Fisches, Zacharias V. and Paliskara, Sahan and Guo, Simon and Zhang, Alex and Spisak, Joe and Cummins, Chris and Leather, Hugh and Synnaeve, Gabriel and Isaacson, Joe and Markosyan, Aram and Saroufim, Mark},
    year={2025},
    month={6},
    url={https://huggingface.co/facebook/KernelLLM},
}

@inproceedings{GargReinforcementTO,
  title={Reinforcement Tuning Open Source LLMs for Kernel Generation},
  author={Aksh Garg and Jeffrey Heo and Megan Mou},
  url={https://api.semanticscholar.org/CorpusID:281103975}
}

@article{woo2025tritonrl,
  title={Tritonrl: Training llms to think and code triton without cheating},
  author={Woo, Jiin and Zhu, Shaowei and Nie, Allen and Jia, Zhen and Wang, Yida and Park, Youngsuk},
  journal={arXiv preprint arXiv:2510.17891},
  year={2025}
}

@article{hahn2023theory,
  title={A theory of emergent in-context learning as implicit structure induction},
  author={Hahn, Michael and Goyal, Navin},
  journal={arXiv preprint arXiv:2303.07971},
  year={2023}
}

@inproceedings{wu2025context,
  title={Why In-Context Learning Models are Good Few-Shot Learners?},
  author={Wu, Shiguang and Wang, Yaqing and Yao, Quanming},
  booktitle={The Thirteenth International Conference on Learning Representations},
  year={2025}
}

@article{dai2026cuda,
  title={Cuda agent: Large-scale agentic rl for high-performance cuda kernel generation},
  author={Dai, Weinan and Wu, Hanlin and Yu, Qiying and Gao, Huan-ang and Li, Jiahao and Jiang, Chengquan and Lou, Weiqiang and Song, Yufan and Yu, Hongli and Chen, Jiaze and others},
  journal={arXiv preprint arXiv:2602.24286},
  year={2026}
}

@article{baronio2025kevin,
  title={Kevin: Multi-turn rl for generating cuda kernels},
  author={Baronio, Carlo and Marsella, Pietro and Pan, Ben and Guo, Simon and Alberti, Silas},
  journal={arXiv preprint arXiv:2507.11948},
  year={2025}
}

@article{li2025cuda,
  title={Cuda-l1: Improving cuda optimization via contrastive reinforcement learning},
  author={Li, Xiaoya and Sun, Xiaofei and Wang, Albert and Li, Jiwei and Shum, Chris},
  journal={arXiv preprint arXiv:2507.14111},
  year={2025}
}

@article{dong2026kernelblaster,
  title={KernelBlaster: Continual Cross-Task CUDA Optimization via Memory-Augmented In-Context Reinforcement Learning},
  author={Dong, Kris Shengjun and Modi, Sahil and Nikiforov, Dima and Damani, Sana and Lin, Edward and Hari, Siva Kumar Sastry and Kozyrakis, Christos},
  journal={arXiv preprint arXiv:2602.14293},
  year={2026}
}

@article{dong2025stark,
  title={Stark: Strategic team of agents for refining kernels},
  author={Dong, Juncheng and Yang, Yang and Liu, Tao and Wang, Yang and Qi, Feng and Tarokh, Vahid and Rangadurai, Kaushik and Yang, Shuang},
  journal={arXiv preprint arXiv:2510.16996},
  year={2025}
}

@article{zhang2025cudaforge,
  title={Cudaforge: An agent framework with hardware feedback for cuda kernel optimization},
  author={Zhang, Zijian and Wang, Rong and Li, Shiyang and Luo, Yuebo and Hong, Mingyi and Ding, Caiwen},
  journal={arXiv preprint arXiv:2511.01884},
  year={2025}
}

@article{xu2024does,
  title={Does few-shot learning help LLM performance in code synthesis?},
  author={Xu, Derek and Xie, Tong and Xia, Botao and Li, Haoyu and Bai, Yunsheng and Sun, Yizhou and Wang, Wei},
  journal={arXiv preprint arXiv:2412.02906},
  year={2024}
}

@article{chen2026avo,
  title={AVO: agentic variation operators for autonomous evolutionary search},
  author={Chen, Terry and Ye, Zhifan and Xu, Bing and Ye, Zihao and Liu, Timmy and Hassani, Ali and Chen, Tianqi and Kerr, Andrew and Wu, Haicheng and Xu, Yang and others},
  journal={arXiv preprint arXiv:2603.24517},
  year={2026}
}

@article{wen2025multikernelbench,
  title={Multikernelbench: A multi-platform benchmark for kernel generation},
  author={Wen, Zhongzhen and Zhang, Yinghui and Li, Zhong and Liu, Zhongxin and Xie, Linna and Zhang, Tian},
  journal={arXiv e-prints, pp. arXiv--2507},
  year={2025}
}

@inproceedings{liu2025few,
  title={Few-shot natural language to first-order logic translation via code generation},
  author={Liu, Junnan},
  booktitle={Proceedings of the 2025 Conference of the Nations of the Americas Chapter of the Association for Computational Linguistics: Human Language Technologies (Volume 1: Long Papers)},
  pages={10939--10960},
  year={2025}
}

@article{zhang2025qwen3,
  title={Qwen3 embedding: Advancing text embedding and reranking through foundation models},
  author={Zhang, Yanzhao and Li, Mingxin and Long, Dingkun and Zhang, Xin and Lin, Huan and Yang, Baosong and Xie, Pengjun and Yang, An and Liu, Dayiheng and Lin, Junyang and others},
  journal={arXiv preprint arXiv:2506.05176},
  year={2025}
}

@inproceedings{samuel2025mmmorrf,
  title={Mmmorrf: Multimodal multilingual modularized reciprocal rank fusion},
  author={Samuel, Saron and DeGenaro, Dan and Guallar-Blasco, Jimena and Sanders, Kate and Eisape, Seun and Reddy, Arun and Martin, Alexander and Yates, Andrew and Yang, Eugene and Carpenter, Cameron and others},
  booktitle={Proceedings of the 48th International ACM SIGIR Conference on Research and Development in Information Retrieval},
  pages={4004--4009},
  year={2025}
}

@article{rackauckas2024rag,
  title={Rag-fusion: a new take on retrieval-augmented generation},
  author={Rackauckas, Zackary},
  journal={arXiv preprint arXiv:2402.03367},
  year={2024}
}

@online{cannbot,
  title        = {CANNBot},
  author       = {Huawei},
  year         = 2026,
  url          = {https://gitcode.com/cann/cannbot-skills},
  urldate      = {2026-6-30}
}

@online{opsnn,
  title        = {ops-nn},
  author       = {Huawei},
  year         = 2026,
  url          = {https://gitcode.com/cann/ops-nn},
  urldate      = {2026-6-30}
}

@article{wang2025faster,
  title={Faster and stronger: Unleashing data processing potential through hardware heterogeneity},
  author={Wang, Cong and Luo, Yang and Du, Wenzhuo and Wang, Ke and Gu, Naijie and Yu, Jun},
  journal={IEEE Internet of Things Journal},
  volume={12},
  number={10},
  pages={14559--14576},
  year={2025},
  publisher={IEEE}
}

@inproceedings{zhang2025skybyte,
  title={SkyByte: architecting an efficient memory-semantic CXL-based SSD with OS and hardware co-design},
  author={Zhang, Haoyang and Xue, Yuqi and Zhou, Yirui Eric and Li, Shaobo and Huang, Jian},
  booktitle={2025 IEEE International Symposium on High Performance Computer Architecture (HPCA)},
  pages={577--593},
  year={2025},
  organization={IEEE}
}

@inproceedings{fang2026heterogeneous,
  title={Heterogeneous Multi-Computing-Units Based LT Coded Computation in Wireless Networks},
  author={Fang, Borui and Qiao, Xiaorui},
  booktitle={2026 IEEE Wireless Communications and Networking Conference (WCNC)},
  pages={1--6},
  year={2026},
  organization={IEEE}
}

@article{fernando2025edge,
  title={On edge-fog-cloud collaboration and reaping its benefits: a heterogeneous multi-tier edge computing architecture},
  author={Fernando, Niroshinie and Shrestha, Samir and Loke, Seng W and Lee, Kevin},
  journal={Future Internet},
  volume={17},
  number={1},
  pages={22},
  year={2025},
  publisher={MDPI}
}

@article{lagouvardos2025incredible,
  title={The Incredible Shrinking Context... in a decompiler near you},
  author={Lagouvardos, Sifis and Bollanos, Yannis and Grech, Neville and Smaragdakis, Yannis},
  journal={Proceedings of the ACM on Software Engineering},
  volume={2},
  number={ISSTA},
  pages={1350--1373},
  year={2025},
  publisher={ACM New York, NY, USA}
}

@article{shir2025robust,
  title={Robust Vulnerability Detection across Compilations: LLVM-IR vs. Assembly with Transformer Model},
  author={Shir, Rony and Surve, Priyanka and Elovici, Yuval and Shabtai, Asaf},
  journal={Proceedings of the ACM on Software Engineering},
  volume={2},
  number={ISSTA},
  pages={618--639},
  year={2025},
  publisher={ACM New York, NY, USA}
}

@article{zhou2025type,
  title={Type-Alias Analysis: Enabling LLVM IR with Accurate Types},
  author={Zhou, Jinmeng and Pan, Ziyue and Shen, Wenbo and Wang, Xingkai and Lu, Kangjie and Qian, Zhiyun},
  journal={Proceedings of the ACM on Software Engineering},
  volume={2},
  number={ISSTA},
  pages={2203--2226},
  year={2025},
  publisher={ACM New York, NY, USA}
}

@inproceedings{zhan2024react,
  title={React: Ir-level patch presence test for binary},
  author={Zhan, Qi and Hu, Xing and Xia, Xin and Li, Shanping},
  booktitle={Proceedings of the 39th IEEE/ACM International Conference on Automated Software Engineering},
  pages={381--392},
  year={2024}
}

@inproceedings{cormack2009reciprocal,
  title={Reciprocal rank fusion outperforms condorcet and individual rank learning methods},
  author={Cormack, Gordon V and Clarke, Charles LA and Buettcher, Stefan},
  booktitle={Proceedings of the 32nd international ACM SIGIR conference on Research and development in information retrieval},
  pages={758--759},
  year={2009}
}

@inproceedings{askari2023injecting,
  title={Injecting the BM25 score as text improves BERT-based re-rankers},
  author={Askari, Arian and Abolghasemi, Amin and Pasi, Gabriella and Kraaij, Wessel and Verberne, Suzan},
  booktitle={European Conference on Information Retrieval},
  pages={66--83},
  year={2023},
  organization={Springer}
}

@inproceedings{liu2022morest,
  title={Morest: Model-based restful api testing with execution feedback},
  author={Liu, Yi and Li, Yuekang and Deng, Gelei and Liu, Yang and Wan, Ruiyuan and Wu, Runchao and Ji, Dandan and Xu, Shiheng and Bao, Minli},
  booktitle={Proceedings of the 44th International Conference on Software Engineering},
  pages={1406--1417},
  year={2022}
}

@inproceedings{mai2025towards,
  title={Towards Better Answers: Automated Stack Overflow Post Updating},
  author={Mai, Yubo and Gao, Zhipeng and Wang, Haoye and Bi, Tingting and Hu, Xing and Xia, Xin and Sun, Jianling},
  booktitle={2025 IEEE/ACM 47th International Conference on Software Engineering (ICSE)},
  pages={591--603},
  year={2025},
  organization={IEEE}
}

@article{achiam2023gpt,
  title={Gpt-4 technical report},
  author={Achiam, Josh and Adler, Steven and Agarwal, Sandhini and Ahmad, Lama and Akkaya, Ilge and Aleman, Florencia Leoni and Almeida, Diogo and Altenschmidt, Janko and Altman, Sam and Anadkat, Shyamal and others},
  journal={arXiv preprint arXiv:2303.08774},
  year={2023}
}

@article{zeng2026glm,
  title={Glm-5: from vibe coding to agentic engineering},
  author={Zeng, Aohan and Lv, Xin and Hou, Zhenyu and Du, Zhengxiao and Zheng, Qinkai and Chen, Bin and Yin, Da and Ge, Chendi and Huang, Chenghua and Xie, Chengxing and others},
  journal={arXiv preprint arXiv:2602.15763},
  year={2026}
}

@article{team2026qwen3,
  title={Qwen3. 5-omni technical report},
  author={Team, Qwen},
  journal={arXiv preprint arXiv:2604.15804},
  year={2026}
}

@article{xu2026deepseek,
  title={DeepSeek-V4: Towards Highly Efficient Million-Token Context Intelligence},
  author={Xu, Anyi and Lin, Bangcai and Xue, Bing and Wang, Bingxuan and Xu, Bingzheng and Wu, Bochao and Zhang, Bowei and Lin, Chaofan and Dong, Chen and Ling, Chenchen and others},
  journal={arXiv preprint arXiv:2606.19348},
  year={2026}
}

\end{document}